\documentclass[10pt,twocolumn,letterpaper]{article}

\usepackage{cvpr}              

\usepackage[dvipsnames]{xcolor}

\usepackage{multirow}
\usepackage{colortbl}  
\usepackage{xcolor}
\usepackage{array}   
\usepackage{booktabs}
\usepackage{indentfirst}
\usepackage{graphicx}
\usepackage{blindtext}
\usepackage{caption}
\usepackage{subcaption}
\usepackage{ntheorem}

\usepackage[ruled,vlined]{algorithm2e}
\usepackage{stfloats}
\usepackage[accsupp]{axessibility}

\definecolor{cvprblue}{rgb}{0.21,0.49,0.74}
\usepackage[pagebackref,breaklinks,colorlinks,citecolor=cvprblue]{hyperref}
\def\paperID{18} 
\def\confName{CVPR}
\def\confYear{2025}

\title{Domain Generalization through \\Attenuation of Domain-Specific Information}
\author{%
Reiji Saito ~ Kazuhiro Hotta ~ \\
\normalsize
Meijo University \\
\normalsize
{\tt\small 200442065@ccalumni.meijo-u.ac.jp, kazuhotta@meijo-u.ac.jp} 
}

\begin{document}
\maketitle
\begin{abstract}
In this paper, we propose a new evaluation metric called Domain Independence (DI) and Attenuation of Domain-Specific Information (ADSI) which is specifically designed for domain-generalized semantic segmentation in automotive images.
DI measures the presence of domain-specific information: a lower DI value indicates strong domain dependence, while a higher DI value suggests greater domain independence. This makes it roughly where domain-specific information exists and up to which frequency range it is present. As a result, it becomes possible to effectively suppress only the regions in the image that contain domain-specific information, enabling feature extraction independent of the domain. ADSI uses a Butterworth filter to remove the low-frequency components of images that contain inherent domain-specific information such as sensor characteristics and lighting conditions. However, since low-frequency components also contain important information such as color, we should not remove them completely. Thus, a scalar value (ranging from 0 to 1) is multiplied by the low-frequency components to retain essential information. This helps the model learn more domain-independent features. In experiments, GTA5 (synthetic dataset) was used as training images, and a real-world dataset was used for evaluation, and the proposed method outperformed conventional approaches. Similarly, in experiments that the Cityscapes (real-world dataset) was used for training and various environment datasets such as rain and nighttime were used for evaluation, the proposed method demonstrated its robustness under nighttime conditions.
\end{abstract}
\section{Introduction}
\label{sec:intro}
\begin{figure}[t]
\begin{center}
\includegraphics[width=1.0\linewidth]{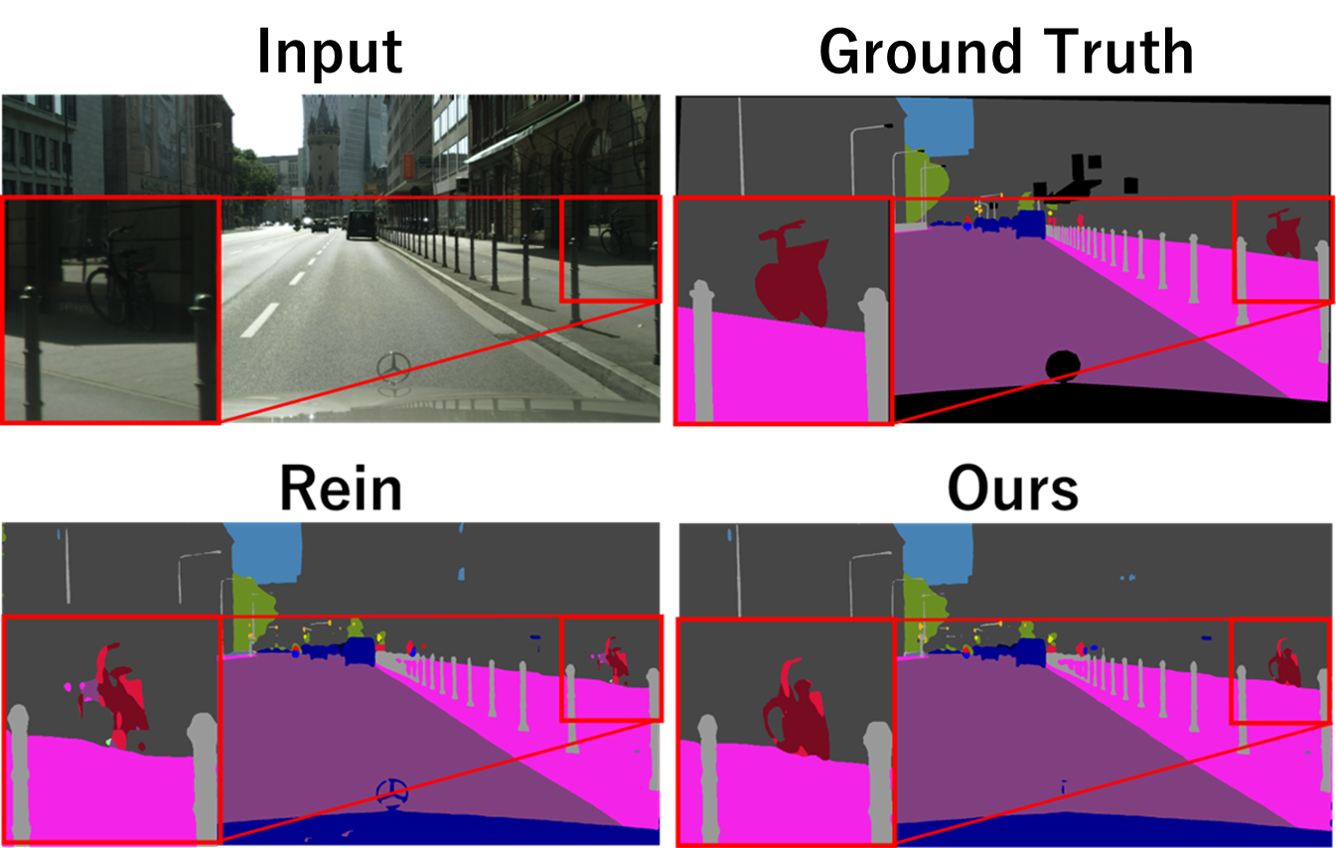}
\end{center}
\caption{The areas improved by the proposed method are highlighted. The red box indicates a bicycle hidden in the shadow, making it quite difficult to see. As a result, the conventional method, Rein, fails to recognize the bicycle. However, by applying the proposed method (Ours), the bicycle hidden in the shadow becomes recognizable.}
\label{fig:visual}
\end{figure}

Domain generalization aims to create models that are robust to unseen images that are not present in the training data, and recent research has achieved impressive results. SIMPLE \cite{simple} achieved the state-of-the-art performance by leveraging multiple pretrained models without fine-tuning and selecting the most suitable pretrained model for each test sample.
Similarly, domain generalization for semantic segmentation (DGSS) of automotive images, which we address in this paper, has improved in performance with the emergence of foundational models. Specifically, Rein \cite{rein} and VLTSeg \cite{VLT} achieved high-performance results by fine-tuning foundation models such as DINOv2 \cite{dino} and vision-language models (VLMs) \cite{clip}.
There are two types of domain generalization: one uses images from multiple domains (e.g. Photo and Sketch), and the other uses images from a single domain. For semantic segmentation, the high cost for obtaining ground-truth annotations often leads to use single-domain images or synthetic data. However, developing highly generalized models using only single-domain images remains challenging and is an active area of research.


The approach for using frequency domain
has proven to be both simple and highly effective for domain adaptation and domain generalization. 
In particular, methods such as FDA \cite{FDA} and amplitude mix \cite{FACT} improved the accuracy by swapping or mixing the low-frequency components of images from multiple domains.
Low-frequency components contain domain-specific information, such as sensor characteristics and lighting differences, so replacing them helps mitigate distribution mismatches. Additionally, the RaffeSDG \cite{raffesdg} method demonstrated generalization performance by attenuating low-frequency components separately for each color channel.

However, there are three issues with these methods.
First, hyperparameters used in the experiment
are determined through visualization or set implicitly, 
making it unclear where domain-specific information exists and to what extent it is considered. As a result, there is a possibility that the settings are overly optimized for the test data.
Second, methods like FDA and amplitude mix assume the use of images from multiple domains. However, in domain generalization for in-vehicle images, training is conducted using only single-domain images, meaning that simply swapping or mixing low-frequency components 
does not necessarily improve generalization. This is because the sensor characteristics and lighting conditions remain consistent.
Third, in the RaffeSDG method, low-frequency components are attenuated differently for each color channel when fundus images are used. However, for the in-vehicle image tasks addressed in this paper, we consider color information to be essential. This is because objects such as grass and sidewalks have similar shapes but are distinguishable by color, and losing this distinction could negatively impact predictions.

To address the first issue, this paper proposes a novel evaluation metric called “Domain Independence (DI)”, which measures the degree of domain dependency. If an image contains domain-specific information, the DI value is low, whereas if it consists only of domain-independent information, the DI value is high.
By using DI, we can roughly determine where domain-specific information is present in the frequency space of an image 
(e.g. it exists in low or high frequencies, in amplitude or phase components).
Furthermore, once the location of domain-specific information is identified by the proposed DI, it becomes possible to realize domain generalization from single domain.

To address the second and third issues, we propose 
“Attenuation of Domain-Specific Information (ADSI)”. ADSI is designed for realizing domain generalization from single domain by using DI metric to guide the hyperparameters. 
DI evaluation revealed that domain-specific information is primarily present in the low-frequency components of images.
Thus, to mitigate domain-specific information, ADSI uses a Butterworth filter to remove the low-frequency components of images that contain domain-specific information. Furthermore, to retain the color information in the low-frequency components that is important for the model's predictions, we multiply the Butterworth filter by a randomly chosen scalar value between 0 and 1.
Additionally, to maintain color consistency, ADSI applies the same scalar value to all color channels rather than using different values for each channel.
This approach enables learning with domain-independent information, improving the robustness to unseen domains and enhancing generalization performance.

In experiments, we used GTA5 \cite{gta}, which is a synthetic dataset, as training data and evaluated the generalization performance on real-world datasets such as Cityscapes \cite{city}, BDD \cite{bdd}, and Mapillary \cite{map}. Additionally, we conducted experiments using Cityscapes as the training dataset and evaluated performance on ACDC \cite{acdc}, which includes challenging conditions such as rain and nighttime scenes.
For comparison, we used Rein \cite{rein}, a state-of-the-art domain generalization model known for its simplicity and high accuracy. As a result, in the experiments where GTA5 was used as training data, ADSI achieved an average mIoU improvement of 1.39$\%$ across all datasets compared to Rein. 
Furthermore, in the experiments where Cityscapes was used as training data, ADSI outperformed Rein by 3.26$\%$ mIoU on nighttime data, demonstrating its effectiveness in challenging conditions.

The main contributions of this paper are as follows.
\begin{itemize}
    \item We propose “Domain Independence (DI)”. DI clarifies the regions containing domain-specific information and enables the quantitative setting of experimental parameters.
    \item We propose “Attenuation of Domain-Specific Information (ADSI)”, which can realize domain generalization from single domain.
    \item We conducted experiments across various DGSS tasks and demonstrated that the proposed ADSI achieved superior generalization performance compared to the state-of-the-art methods. Additionally, it was shown to outperform conventional methods, such as amplitude mix and RaffeSDG.
\end{itemize}

The structure of this paper is as follows. In \cref{sec:2_related}, we discuss related works. \cref{sec:3_proposed} explains the details of the proposed method. In \cref{sec:4_experiments}, we present experimental results and discussion. Finally, \cref{sec:5_conclusion} concludes our paper and describes future challenges.

\section{Related Works}
\label{sec:2_related}

\textbf{Domain Generalization for Semantic Segmentation (DGSS)} aims to build a robust model that can handle unknown domains not included in the training images. Unlike domain adaptation (DA), which focuses on a single target domain, DGSS requires the ability to generalize to any datasets. Previous methods have included normalization techniques, whitening approaches, domain randomization, and strategies utilizing foundation models.
For example, in the normalization-based approach IBN \cite{ibn}, instance normalization is used to learn domain-invariant features, while batch normalization is used to learn content-specific information. By carefully integrating these two types of normalization, IBN improved generalization capability. In RobustNet \cite{robust} which is the whitening-based approach, features that have changed due to optical transformations are selectively removed, thereby erasing domain-specific style information. With domain randomization, WildNet \cite{wild} diversifies style by making the source domain more similar to the style in the ImageNet \cite{image} dataset. Finally, in methods based on foundation models, Rein examines the generalization performance of models such as CLIP \cite{clip} and DINOv2 \cite{dino}, and it improved their generalization ability through fine-tuning.

However, these methods do not clarify where domain-specific information resides in the image or how it leads to performance degradation. Therefore, in this paper, we use the GTA5 and ImageNet datasets to identify which parts of an image contain domain-specific information. Moreover, inspired by RobustNet which selectively removes changes caused by optical properties, we use a Butterworth filter to remove low-frequency components that contain domain-specific information. However, since low-frequency components also include critical information for predictions, such as color information, we multiply them by a scalar value between 0 and 1, ensuring they are not entirely removed but effectively utilized.

\section{Methodology}
\label{sec:3_proposed}

In this paper, we aim to improve the accuracy of DGSS. First, in \cref{sub:di}, we propose an evaluation metric called “Domain Independence (DI)”. Next, in \cref{sub:QED}, we roughly determine which parts of an image contain domain-specific information. Finally, in \cref{sub:ADSI}, we propose an image transformation method for in-vehicle image DGSS, called “Attenuation of Domain-Specific Information (ADSI)”.

\subsection{Domain Independence (DI)}
\label{sub:di}

\begin{figure}[t]
\begin{center}
\includegraphics[width=1.0\linewidth]{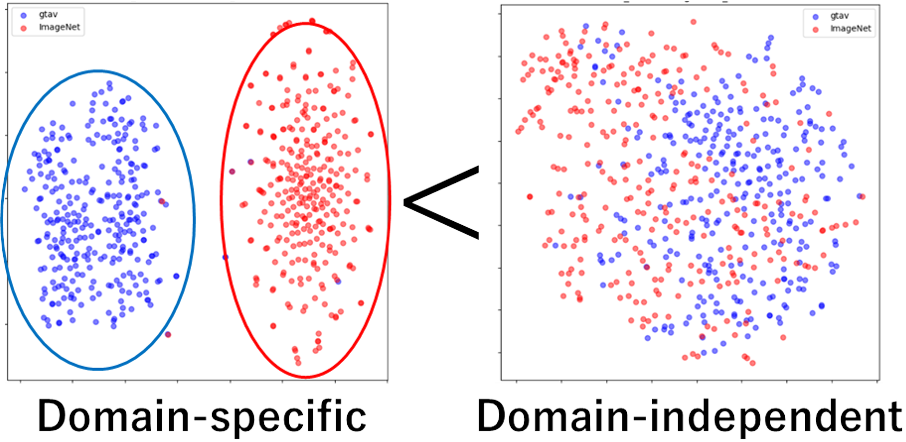}
\end{center}
\caption{Definition of Domain-Specific and Domain-Independent. If the data has specific features to a particular domain, Domain-Independent metric will be low. On the other hand, if it does not contain domain-specific information, Domain-Independent metric will be high.}
\label{fig:sp_in}
\end{figure}

In this paper, we roughly determine which parts of an image contain domain-specific information. First, \cref{fig:sp_in} explains what domain-specific information is. In \cref{fig:sp_in}, two datasets (indicated in red and blue) are fed into the model, and the output features are visualized using t-SNE. Domain-specific information refers to the case where the two datasets are clustered separately, as shown on the left side of the figure. On the other hand, domain-independent information refers to the case where the two datasets are mixed. In other words, when the feature distances between different datasets are close, it indicates domain-independent information.

\begin{figure}[t]
\begin{center}
\includegraphics[width=1.0\linewidth]{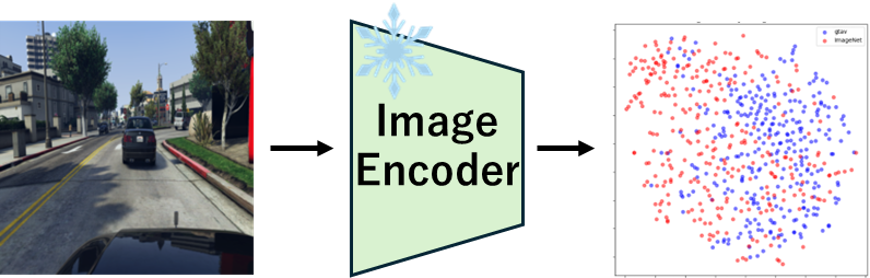}
\end{center}
\caption{Overview of Domain Independence metric.
The image is passed through a frozen Image Encoder to extract its features. The reason the image encoder is frozen is that it ensures consistent feature extraction, allowing roughly evaluation of differences between image domains. After that, by comparing the distances between features, the degree to which the image depends on a specific domain can be quantified.
}
\label{fig:di}
\end{figure}

\begin{figure}[t]
\begin{center}
\includegraphics[width=1.0\linewidth]{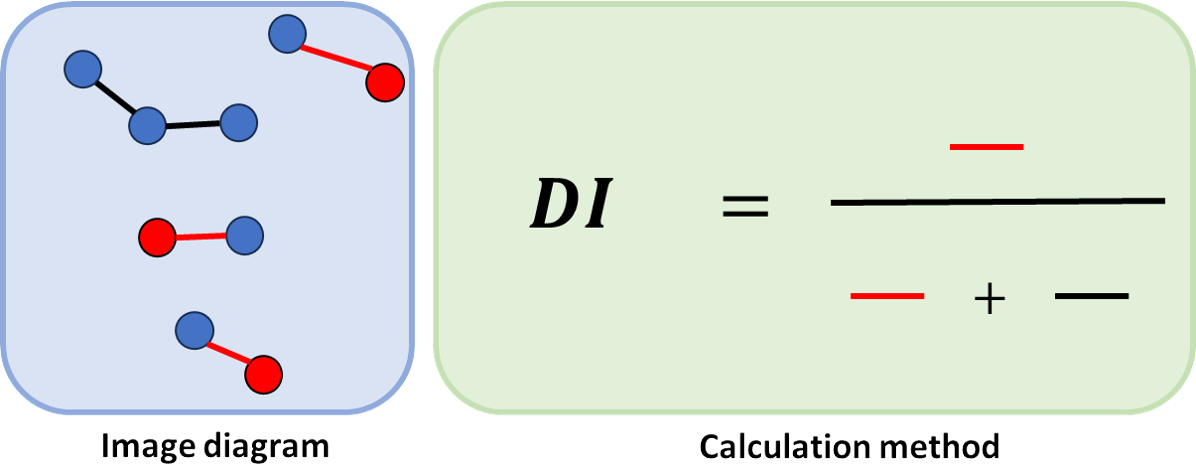}
\end{center}
\caption{DI image diagram and calculation method. For a certain feature, if the closest feature values are in the same dataset, we count the black lines. On the other hand, if they are in different datasets, count the red lines.}
\label{fig:di_cal}
\end{figure}

Following this definition, we roughly determine where domain-specific information is present in the frequency space of images using the GTA5 dataset as the training data and the ImageNet dataset. We use ImageNet because it is a dataset of natural images with diverse styles, making it suitable for covering a wide range of image variations. \cref{fig:di} provides an overview of “Domain Independence (DI)”.

First, we define the input image $x\in \mathbb{R}^{3\times H \times W}$, where $H$ and $W$ represent the height and width, respectively. This image is fed into a pre-trained Vision Transformer (ViT) \cite{ViT}, and we obtain the output feature map $f\in \mathbb{R}^{N\times D}$ where $N$ and $D$ denote the number of patches and the embedding dimension, respectively. We extract $B$ feature maps from the GTA5 and ImageNet datasets,
resulting in a total of $2B$ feature maps. In this study, we set 
$B = 300$.

Next, we compute the pairwise distance matrix $P\in \mathbb{R}^{2B\times 2B}$. Excluding the diagonal elements of $P$, we determine the closest distance for each feature map. We assess the degree of domain independence based on whether the nearest feature map belongs to the same dataset or a different one. Specifically, for each feature map, if the nearest feature map is from the same dataset, we increment the denominator by 1. If it is from a different dataset, we increment both the numerator and the denominator by 1. 
\begin{eqnarray}
  DI &=& \frac{M}{M + N}
  \label{equation:domain}
\end{eqnarray}
where $N$ and $M$ are the counts of these occurrences, respectively.

The overview of \cref{equation:domain} is illustrated in \cref{fig:di_cal}. A higher DI value indicates that the feature does not contain domain-specific information, while a lower DI value suggests the presence of domain-specific information.

\subsection{Quantitative Evaluation of Domain-Specific Information}
\label{sub:QED}

In this paper, we use DI to determine roughly which regions in the frequency space contain domain-specific information. The reason for analyzing the frequency space is that low-frequency amplitude components are believed to contain domain-specific information \cite{FDA}. Therefore, we progressively remove low-frequency components to determine up to which frequency domain-specific information is retained. Additionally, we clarify whether the amplitude component, the phase component, or both in the frequency space contain domain-specific information.

\subsubsection{Experiment on Removing Low-Frequency Components}
\label{subsub:low}

\begin{figure}[t]
    \begin{center}
    \includegraphics[width=1.0\linewidth]{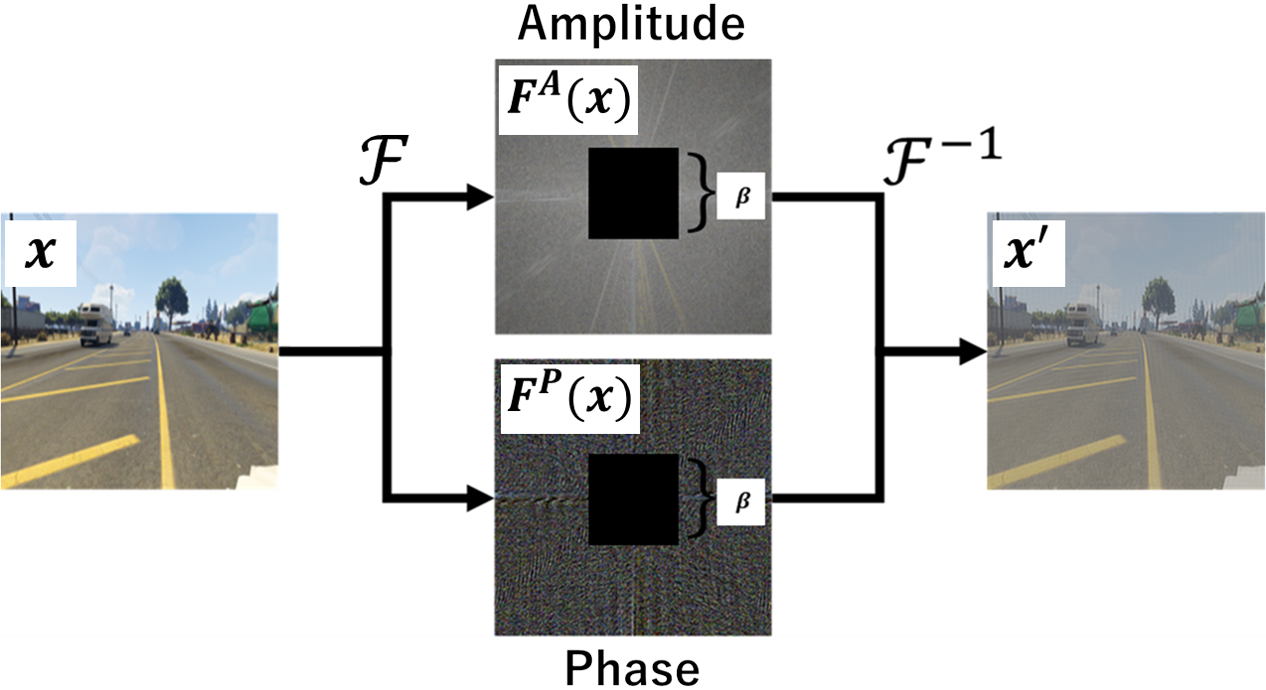}
    \end{center}
    \caption{Overview of low-frequency component removal. By removing the frequency components inside the box, 
    sensor characteristics and lighting conditions
    present in the low-frequency components can be eliminated. Since filters like the Butterworth filter gradually reduce the frequency components, it is not clear where the domain-specific information is contained. Therefore, here we use a box filter to clearly eliminate the frequency components in specific areas. Additionally, by adjusting $\beta$, the size of the box can be modified. 
    }
    \label{fig:zikken}
\end{figure}

We remove the low-frequency components of an image using a box and quantitatively evaluate the size of the box that contains domain-specific information using a domain information index (DI). First, we prepare an input image $x \in \mathbb{R}^{3\times H \times W}$ and apply the Fast Fourier Transform (FFT) to obtain the frequency representation $F(x)$, which consists of amplitude and phase components.
\begin{eqnarray}
\mathcal{F}(x)(m,n) = \sum_{h,w} x(h,w) \ e^{-j 2\pi\!\left(\frac{h}{H}m + \frac{w}{W}n\right)},\; j^2 = -1
\label{eq:FFT}
\end{eqnarray}
where $(h, w)$ represents the spatial coordinates of the image, while $(m, n)$ represents the coordinates in the frequency domain.

Next, we apply a Box filter and create a mask where the inside of the Box is set to 0 and the outside is set to 1.
\begin{eqnarray}
M_{\beta}(h, w) = 1 - {1}_{(h,w) \in [-\beta H : \beta H, -\beta W : \beta W]}
\label{eq:maskb}
\end{eqnarray}
where $\beta \in (0, 1)$, and the center of the image is assumed to be $(0,0)$. The specified low-frequency components are removed by multiplying the mask with the amplitude and phase components of $F(x)$. Then, an inverse Fourier transform (IFFT) is applied to reconstruct the image with the low-frequency components removed. These processes are illustrated in \cref{fig:zikken}.
\begin{eqnarray}
    x^{\prime} = \mathcal{F}^{-1}  \left[ M_{\beta} \circ \mathcal{F}^{A}(x) 
    , M_{\beta} \circ \mathcal{F}^{P}(x) \right]
\label{eq:iFFT}
\end{eqnarray}
where $\mathcal{F}^{A}$ and $\mathcal{F}^{P}$ represent the amplitude and phase components of the Fourier transform, respectively, while $\mathcal{F}^{-1}$ denotes the inverse Fourier transform. The reconstructed image after the inverse Fourier transform is denoted as $x^{\prime}$.

\begin{figure}[t]
    \begin{center}
    \includegraphics[width=1.0\linewidth]{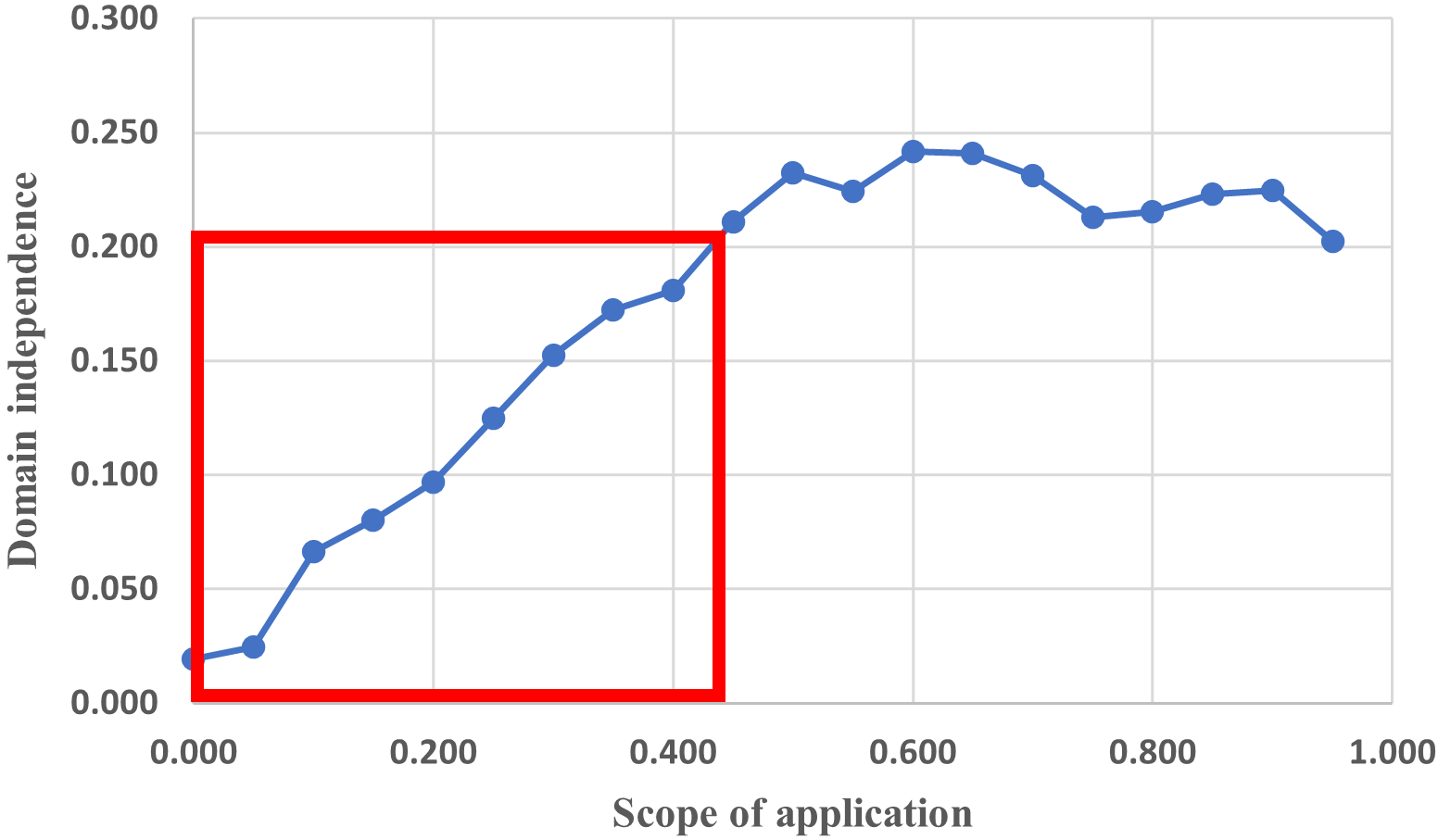}
    \end{center}
    \caption{An experiment was conducted to quantitatively evaluate domain independence using the proposed DI, up to which frequency band an image retains its domain-specific characteristics. 
    The areas enclosed by the red box indicate the regions that exhibit domain-specific characteristics.}
    \label{fig:DI_vi}
\end{figure}

We analyze the influence of low-frequency components by performing a quantitative evaluation using DI on the reconstructed images. Furthermore, by gradually increasing the Box size $(\beta)$, we can determine up to which frequency components retain domain-specific information.
\cref{fig:DI_vi} illustrates an experiment in which DI is used to quantitatively evaluate up to which frequency band an image retains its domain-specific characteristics. The horizontal axis represents the ratio $(\beta)$ of the box size to the image size, while the vertical axis indicates the proposed DI value. Observing \cref{fig:DI_vi}, we can see that the DI value increases from 0 to 0.4 on the horizontal axis. However, when the horizontal axis value exceeds 0.4, the DI value becomes nearly constant. These results suggest that the frequency band up to 0.4 times the image size retains domain-specific characteristics, and applying processing to this region can effectively mitigate domain-specific information. Therefore, in this experiment, the value of $\beta$ is set between 0 and 0.4.

\subsubsection{Comparison of Domain Independence Between Amplitude and Phase Components}
\label{subsub:am_pha}

Conventional methods such as FDA and amplitude mix apply Fourier transform to images, process only the amplitude components, and then reconstruct the images for training. The phase components were not processed because they were considered to contain high-level semantics and were believed to be unaffected by domain shifts. However, does domain shift truly not occur? To investigate this, we conduct a quantitative evaluation using Domain Invariance (DI) with images reconstructed from only the amplitude components and only the phase components.

We conducted an evaluation using Domain Invariance (DI) with images reconstructed from only the amplitude components and only the phase components. DI values were 0.19 for the amplitude-only reconstruction and 0.10 for the phase-only reconstruction.
From these results, it is confirmed that the phase components contain intrinsic domain-specific information. Therefore, by applying a filter to both the amplitude and phase components, we can mitigate domain-specific information.

\subsection{Attenuation of Domain-Specific Information (ADSI)}
\label{sub:ADSI}

\begin{figure}[t]
    \begin{center}
    \includegraphics[width=1.02\linewidth]{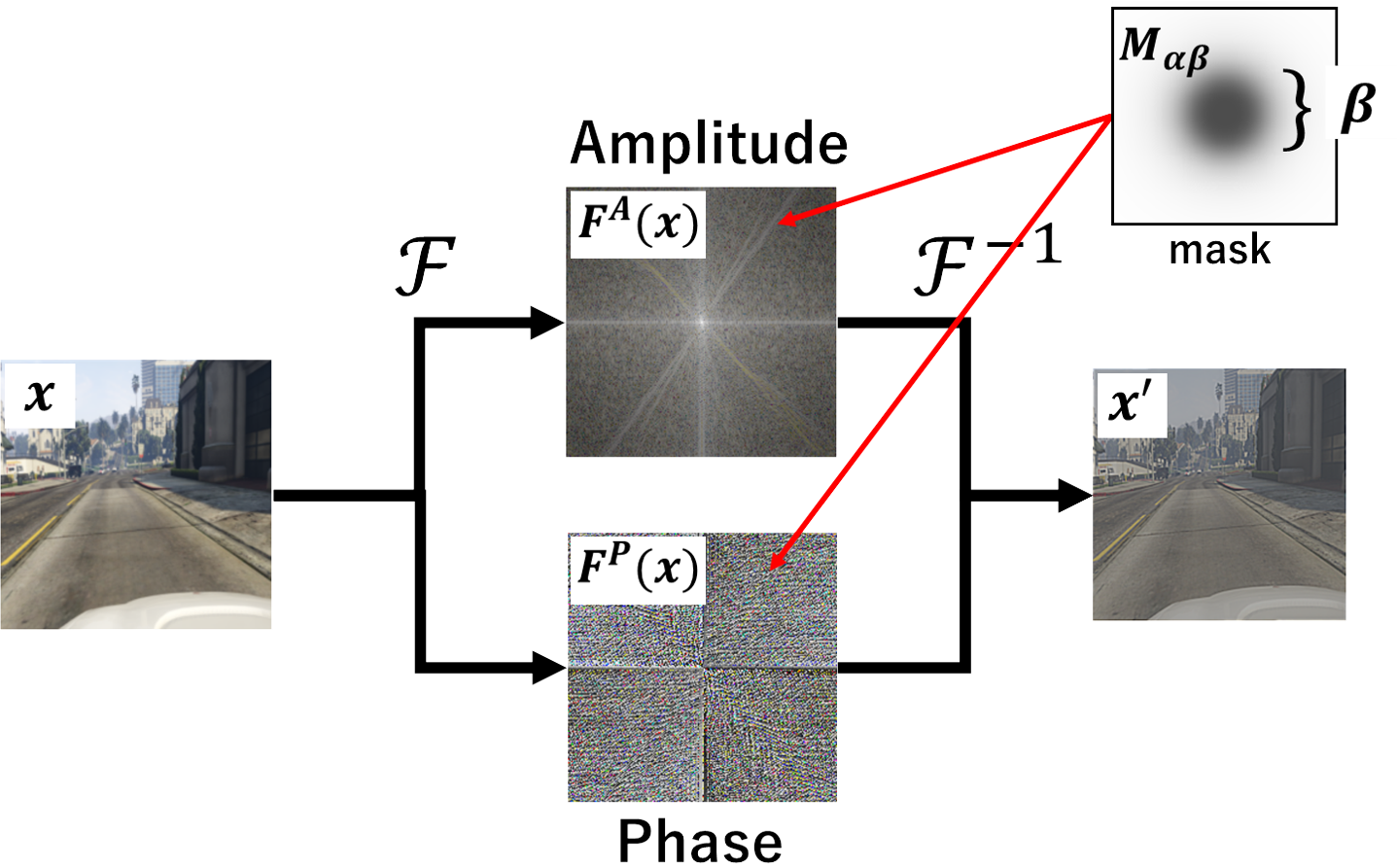}
    \end{center}
    \caption{Overview of the Proposed Method. The proposed method applies a Fourier transform to the input image and multiplies both the amplitude and phase components by $M_b$. This process removes low-frequency components that contain domain-specific information, such as sensor characteristics. Then, the image is reconstructed using the inverse Fourier transform. By training with these reconstructed images, the model can make predictions that are independent of domain-specific information.}
    \label{fig:pro}
\end{figure}

We propose “Attenuation of Domain-Specific Information (ADSI)”, in which domain generalization is obtained from a single domain. \cref{fig:pro} illustrates the overview of ADSI. First, we define the input image $x$ and perform a fast Fourier transform using Eq. \ref{eq:FFT}. Next, as shown in \cref{subsub:low,subsub:am_pha}, domain-specific information exists in both the amplitude and phase components within the frequency band where $\beta$ is up to 0.4. Therefore, to suppress the influence of low-frequency components in both the amplitude and phase components, we create a mask. 
However, unlike $M_{\beta}$, we use a Butterworth filter. This is because ringing artifacts appear in the output image $x^\prime$ in \cref{fig:zikken}. 
Regarding the consistency with the experiment in \cref{fig:DI_vi}, it differs slightly when a Box is used, but the overall aspects remain unchanged.
The cause of the ringing artifacts is the gap that occurs when the low-frequency components are removed by the Box filter. The mask using a Butterworth filter can be expressed by the following equation.
\begin{eqnarray} 
M_{\beta}^{\prime} = \frac{1}{1 + \left( \frac{\sqrt{h^2 + w^2}}{\beta + \epsilon} \right)^{2N}}
\end{eqnarray}
where the center coordinate refers to $(0,0)$, $\beta \in (0,0.4)$, $N \in (1, 3)$ same as RaffeSDG \cite{raffesdg}
and $\epsilon$ represents a small value.
$M_{\beta}^{\prime}$ is close to 1 for low-frequency components and close to 0 for high-frequency components. In segmentation using in-vehicle images, color information is also necessary. Therefore, to avoid removing low-frequency components completely that contain color information, a random scalar value between 0 and 1 is multiplied by the mask. Then, this mask retains the low-frequency components and removes the high-frequency components. By subtracting 1, the high-frequency components are preserved, and the influence of the low-frequency components is reduced.
The mask can be expressed by the following equation.
\begin{eqnarray} %
M_{\alpha\beta} = 1 - \alpha \circ M_{\beta}^{\prime}
\end{eqnarray}
where $\alpha \in (0, 1)$. By using Eq. \ref{eq:iFFT}, $M_{\alpha\beta}$ is applied to the amplitude and phase components, and finally an inverse Fourier transform is performed to obtain the output image $x^{\prime}$. By using the output image, learning can be conducted with domain-independent information, enabling robust predictions even for unseen data.

\section{Experiments}
\label{sec:4_experiments}

\subsection{Settings}

\textbf{Datasets.} We evaluate the proposed method and existing methods using synthetic datasets (GTA5 \cite{gta}, Synthia \cite{syn}) and real-world datasets (Cityscapes \cite{city} (Citys), BDD100K \cite{bdd} (BDD), Mapillary \cite{map} (Map), ACDC \cite{acdc}). The GTA5 dataset consists of $24,966$ synthetic images obtained from a game with pixel-level annotations. The Synthia dataset contains $9,400$ annotated images rendered from a virtual city. The Cityscapes dataset is an autonomous driving dataset that includes $2,975$ training images and $500$ validation images. Similarly, the BDD100K and Mapillary datasets provide $1,000$ and $2,000$ validation image for perception algorithms in autonomous driving. The ACDC dataset includes challenging conditions such as fog, nighttime, rain, and snow, with validation data consisting of $106$ images for nighttime and $100$ images for each of the other conditions. In addition, to ensure consistency in training data, all datasets are aligned to match the Cityscapes $19$ class format.

\textbf{Implementation details.}
We conduct experiments using Rein \cite{rein} as the baseline. Rein is a state-of-the-art method that utilizes various image foundation models, such as CLIP \cite{clip} and SAM \cite{sam}, as the backbone and employs Mask2Former \cite{mask2former} for the segmentation head. Therefore, we will use Rein (with DINOv2 \cite{dino} as the backbone) as the baseline and evaluate the performance when we incorporate the proposed method into the baseline. The experimental setup follows that of Rein, using AdamW \cite{adamw} as the optimization method. The learning rate is set to $1 \times e^{-5}$ for the backbone and $1 \times e^{-4}$ for Rein and the segmentation head. Training is conducted for $40,000$ iterations with a batch size of $4$, and images are cropped to $512 \times 512$. 
Data augmentation follows the same approach as Rein with ADSI
when the proposed method is used. The evaluation metric is Intersection over Union (IoU). The values presented in Tables represent the mean IoU. The experiments were conducted using a single RTX A6000 GPU for training. For comparison with the baseline, we used DINOv2-Large.

\subsection{Quantitative Evaluation}
\begin{table}[t]
\caption{Comparison with state-of-the-art methods. The model was trained on the GTA5 dataset and evaluated on real-world dataset. Additionally, Avg. represents the average value across Cityscapes (Citys), BDD100K (BDD), and Mapillary (Map).}
\centering
\scalebox{0.77}{ 
\begin{tabular}{l c r c c c c}
    \hline
    \multirow{2}{*}{Method} & \multirow{2}{*}{Backbone} & \multicolumn{4}{c}{GTA5-to-Real} \\ 
    \cline{3-6}
    & & Citys & BDD & Map & Avg. \\
    \hline
    IBN \cite{ibn} & RN50 \cite{resnet} & 33.61 & 32.67 & 37.43 & \cellcolor{gray!10}34.57 \\
    ISW \cite{robust} & RN50 & 36.40 & 35.79 & 40.30 &  \cellcolor{gray!10}37.50\\
    WildNet \cite{wild} & RN101 & 44.62 & 38.35 & 46.09 & \cellcolor{gray!10}43.02 \\
    \hline
    DAFormer \cite{daformer} & MiT-B5 \cite{Segformer} & 52.88 & 47.37 & 54.34 & \cellcolor{gray!10}51.53 \\
    HRDA \cite{hrda} & MiT-B5 & 57.45 & 49.10 & 59.92 & \cellcolor{gray!10}55.49 \\
    VLTSeg \cite{VLT} & EVA02-CLIP-L \cite{eva02} & 65.30 & 58.30 & 66.0 & \cellcolor{gray!10}63.20 \\    
    \hline
    Rein \cite{rein} & CLIP-L \cite{clip} & 58.35 & 54.75 & 59.91 & \cellcolor{gray!10}57.67 \\
    Rein & SAM-H \cite{sam} & 59.66 & 50.03 & 61.14 & \cellcolor{gray!10}54.94 \\
    Rein & EVA02-CLIP-L & 63.23 & 59.63 & 63.81 & \cellcolor{gray!10}62.22 \\
    Rein & DINOv2-L \cite{dino} & 66.12 & 60.51 & 65.90 & \cellcolor{gray!10}64.18 \\
    \hline
    RaffeSDG \cite{raffesdg} & DINOv2-L & 61.78 & 58.98 & 62.72 & \cellcolor{gray!10}61.16 \\
    FACT \cite{FACT} & DINOv2-L & 64.75 & 60.27 & 64.66 & \cellcolor{gray!10}63.23 \\
    \hline
    \cellcolor{blue!10}Ours & \cellcolor{blue!10}DINOv2-L & \cellcolor{blue!10}\textbf{67.75} & \cellcolor{blue!10}\textbf{61.38} & \cellcolor{blue!10}\textbf{67.59} & \cellcolor{blue!10}\textbf{65.57} \\
    \hline
\end{tabular} 
}
\label{table:one}
\end{table}

\cref{table:one} presents the results of training on the synthetic dataset GTA5 and evaluating on the real-world datasets Citys, BDD, and Map. The baseline method, Rein-DINOv2-L, achieves better accuracy compared to other conventional approaches.
Additionally, the conventional methods such as RaffeSDG and FACT were trained using Rein-DINOv2-L but resulted in lower accuracy. This decline in performance may be attributed to the removal or alteration of color information, which negatively impacted predictions.
On the other hand, the proposed method (Ours) achieved accuracy improvements of 2.63$\%$ on Citys, 0.87$\%$ on BDD, and 1.69$\%$ on Map compared to the baseline Rein-DINOv2-L. This demonsrated that our method successfully learned domain-independent features while preserving color information, leading to better generalization performance on real-world data.

\begin{table}[t]
\caption{Comparison with the baseline (Rein) when we trained on Cityscapes (Citys) and evaluated on the ACDC dataset which contains data from various environmental conditions.}
\centering
\scalebox{0.85}{
\begin{tabular}{l c r c c c c}
    \hline
    \multirow{2}{*}{Method} & \multirow{2}{*}{Backbone} & \multicolumn{4}{c}{Citys-to-ACDC} \\ 
    \cline{3-7}
    & & Night & Snow & Fog & Rain & Avg. \\
    \hline
    Rein & DINOv2-L & 55.56 & \textbf{72.13} & 81.09 & \textbf{75.19} & \cellcolor{gray!10}\textbf{70.99} \\
    \cellcolor{blue!10}Ours & \cellcolor{blue!10}DINOv2-L & \cellcolor{blue!10}\textbf{58.82} & \cellcolor{blue!10}71.07 & \cellcolor{blue!10}\textbf{81.17} & \cellcolor{blue!10}69.19 & \cellcolor{blue!10}70.21\\
    \hline
\end{tabular} 
}
\label{table:two}
\end{table}

\cref{table:two} evaluates whether the model trained on Citys generalizes well across various environmental conditions. Since Rein-DINOv2-L achieved the best accuracy among conventional domain generalization methods in Table 1, we compare only with Rein-DINOv2-L.
The results show that Ours improved the accuracy by 3.26$\%$ in the Night scenario compared to Rein-DINOv2-L. This significant improvement in Night conditions may be due to reducing the influence of low-frequency components, which helps the model better capture nighttime environments.
However, in other conditions, the accuracy either remained nearly the same or worsened. This may be because reducing low-frequency components led to a prediction bias toward high-frequency features. 
Since high-frequency components contain the details such as fog or raindrops, and the model was not explicitly trained to handle such domain shifts, this likely resulted in the observed performance drop.

\begin{table}[t]
\caption{Comparison with the baseline (Rein) when we trained on  synthetic datasets: GTA5 and Synthia, and evaluated on Cityscapes (Citys), BDD100K (BDD), and Mapillary (Map) datasets. The accuracy improved when Synthia was included in comparison with only GTA5 for training. Furthermore, Ours outperformed Rein.}
\centering
\scalebox{0.95}{ 
\begin{tabular}{l c r c c c c}
    \hline
    \multirow{2}{*}{Method} & \multirow{2}{*}{Backbone} & \multicolumn{4}{c}{GTA5+Synthia-to-Real} \\ 
    \cline{3-6}
    & & Citys & BDD & Map & Avg. \\
    \hline
    Rein & DINOv2-L & 66.91 & 60.87 & 67.94 & \cellcolor{gray!10}65.24 \\
    \cellcolor{blue!10}Ours & \cellcolor{blue!10}DINOv2-L & \cellcolor{blue!10}\textbf{69.03} & \cellcolor{blue!10}\textbf{61.76} & \cellcolor{blue!10}\textbf{68.26} & \cellcolor{blue!10}\textbf{66.35} \\
    \hline
\end{tabular} 
}
\label{table:three}
\end{table}

\cref{table:three} presents the results when the Synthia dataset was additionally included in training, building upon the results in \cref{table:one}. We conducted this experiment to verify whether increasing synthetic data further enhances the effectiveness of our method.
The Rein-DINOv2-L in \cref{table:three} achieved better accuracy than the Rein-DINOv2-L in \cref{table:one}. Furthermore, Ours outperformed conventional methods, demonstrating its effectiveness. However, the improvement in accuracy was smaller than the case that only the GTA5 dataset was used for training. These results indicate that our method remains applicable even when additional synthetic data is introduced, reinforcing its robustness.

\begin{table}[t]
\caption{Comparison with the baseline (Rein) when trained on GTA5 and evaluated on the ACDC dataset, which contains data from various environmental conditions.}
\centering
\scalebox{0.85}{ 
\begin{tabular}{l c r c c c c}
    \hline
    \multirow{2}{*}{Method} & \multirow{2}{*}{Backbone} & \multicolumn{4}{c}{GTA5-to-ACDC} \\ 
    \cline{3-7}
    & & Night & Snow & Fog & Rain & Avg. \\
    \hline
    Rein & DINOv2-L & 49.40 & 64.74 & 72.19 & 61.63 & \cellcolor{gray!10}61.99 \\
    \cellcolor{blue!10}Ours & \cellcolor{blue!10}DINOv2-L & \cellcolor{blue!10}\textbf{50.37} & \cellcolor{blue!10}\textbf{66.69} & \cellcolor{blue!10}\textbf{72.44} & \cellcolor{blue!10}\textbf{64.57} & \cellcolor{blue!10}\textbf{63.52}\\
    \hline
\end{tabular} 
}
\label{table:four}
\end{table}

\cref{table:four} evaluates whether the model trained on the GTA5 dataset can generalize over various environmental conditions. While \cref{table:two} examined generalization when we trained on real-world data (Citys), our method is trained on synthetic data in this experiment.
The results show that Ours outperforms conventional methods, demonstrating superior accuracy. These findings indicate that the usage of synthetic data leads to performance improvements in all environmental conditions. This improvement may be attributed to our method's ability to remove the characteristic textures of synthetic data, making the model more robust to domain shifts.

\subsection{Qualitative Evaluation}

\begin{figure*}[t]
\begin{center}
\includegraphics[width=1.0\linewidth]{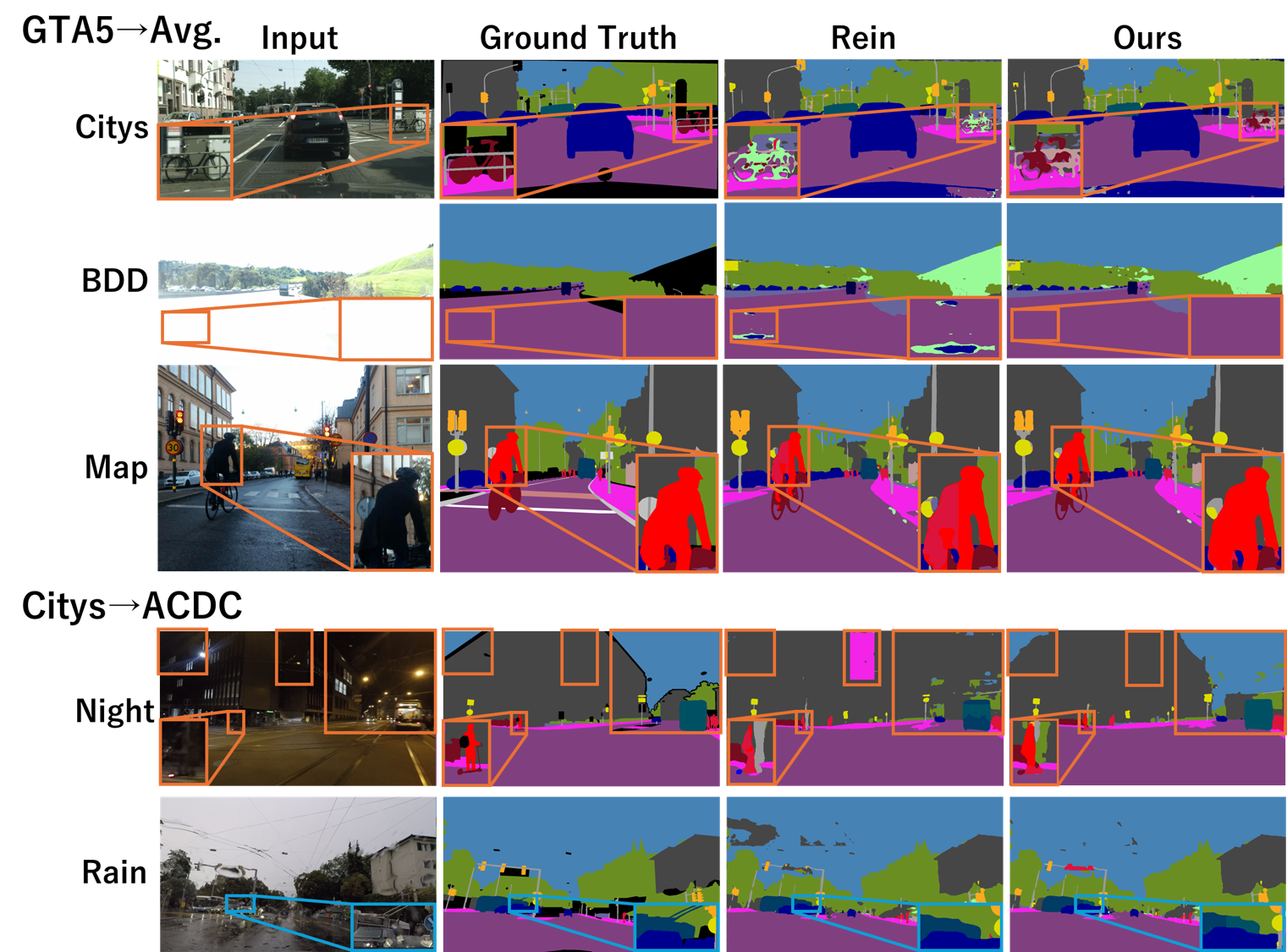}
\end{center}
\caption{The visualization results by baseline model (Rein) and our method (Ours). The results show models trained on GTA5 and evaluated on Cityscapes, BDD, and Map (GTA5 $\rightarrow$ Avg.), as well as models trained on Cityscapes and evaluated on Night and Rain (Cityscapes $\rightarrow$ ACDC).
From left to right, the columns represent Input, Ground Truth, Rein-DINOv2-L, and Ours. The orange boxes highlight improved regions, while the blue boxes indicate degraded areas.}
\label{fig:visual}
\end{figure*}

\cref{fig:visual} shows the visualization results by models trained on GTA5 and evaluated on Cityscapes, BDD, and Map (GTA5 $\rightarrow$ Avg.), as well as models trained on Cityscapes and evaluated on Night and Rain (Cityscapes $\rightarrow$ ACDC).
From left to right, each column represents the input image, ground truth, the baseline model (Rein-DINOv2-L), and our proposed method (Ours). The orange boxes highlight the improved regions, while the blue boxes indicate degraded areas.

In Cityscapes, focusing on the orange boxes, we observe that Rein-DINOv2-L misclassifies the bicycle class as vegetation due to domain shift. In contrast, while it is not perfect, Ours can recognize the bicycle class to some extent. This suggests that our method enables prediction based on domain-independent information.

In Rain, when we focus on the blue boxes, we see that Ours misclassifies the bus as a truck while Rein-DINOv2-L correctly recognizes the bus class. This is likely because Ours focuses on high-frequency components for predictions. However, raindrops in the input image also contain high-frequency components, making it difficult for our model to adapt to the domain shift caused by the training data distribution.

\subsection{Ablation Studies and Analysis}

\begin{table}[t]
\label{table:five}
\caption{The baseline is Rein-DINOv2-L, and the comparison is made when various methods are used for removing low-frequency components. “Box (Amplitude)” refers to the method in \cref{fig:zikken}, where only the amplitude of the low-frequency components is removed by adopting a factor $\alpha$, effectively either removing or keeping the low-frequency components. “Box” follows the same approach as \cref{fig:zikken} but it multiplies $\alpha \in (0,1)$ with $M_b$ of Eq. \ref{eq:maskb} to randomly remove low-frequency components, “Circle” is a variation where “Box” is modified into a Circle. “Color” is created by generating a mask of Ours separately for each color channel and multiplying it with the frequency components. “Butter (Amplitude)” is our proposed method applied only to the amplitude components.}
\centering
\scalebox{0.82}{ 
\begin{tabular}{l c r c c c c}
    \hline
    \multirow{2}{*}{Method} & \multirow{2}{*}{Backbone} & \multicolumn{4}{c}{GTA5-to-Real} \\ 
    \cline{3-6}
    & & Citys & BDD & Map & Avg. \\
    \hline
    Rein & DINOv2-L & 66.12 & 60.51 & 65.90 & \cellcolor{gray!10}64.18 \\
    Box (Amplitude) & DINOv2-L & 66.27 & 61.21 & 66.0 & \cellcolor{gray!10}64.49 \\
    Box & DINOv2-L & 67.19 & 61.14 & 66.22 & \cellcolor{gray!10}64.85 \\
    Circle & DINOv2-L & 67.04 & 60.34 & 66.68 & \cellcolor{gray!10}64.69 \\
    Color & DINOv2-L & 67.39 & 60.08 & 66.66 & \cellcolor{gray!10}64.69 \\
    Butter (Amplitude) & DINOv2-L & 66.84 & 61.56 & 66.50 & \cellcolor{gray!10}64.97 \\
    \hline
    \cellcolor{blue!10}Ours & \cellcolor{blue!10}DINOv2-L & \cellcolor{blue!10}\textbf{67.75} & \cellcolor{blue!10}\textbf{61.38} & \cellcolor{blue!10}\textbf{67.59} & \cellcolor{blue!10}\textbf{65.57} \\
    \hline
\end{tabular} 
}
\end{table}

We evaluate our method by various ablation studies.
The results are shown in \cref{table:five}. Note that the baseline is Rein, and “Box” follows the same approach as \cref{fig:zikken}, where a factor $\alpha$ is multiplied by the low-frequency components to either remove or retain them.
“Circle” is a circular version of Box.
“Color” is created by generating a mask of Ours separately for each color channel and multiplying it with the frequency components.
For methods labeled with “Amplitude”, the adjustment is applied only to the amplitude and not to the phase.

As a result, all methods outperformed the baseline. In addition, Ours achieved the best performance. “Box” and “Circle” did not significantly improve the accuracy due to the presence of ringing artifacts. Furthermore, by applying the augmentation to both amplitude and phase, Ours gave better results than applying it to amplitude alone. This suggests that the phase also has a unique domain, which is consistent with the experimental results in \cref{subsub:am_pha}. Moreover, “Color” applied different masks to each color channel, causing color changes, which likely reduced its effectiveness.

\section{Conclusion}
\label{sec:5_conclusion}

In this paper, we propose DI and ADSI, demonstrating superior results compared to state-of-the-art methods and conventional augmentation techniques. However, our approach did not perform well in rainy and snowy environments durring domain generalization, where the model was trained on Cityscapes and evaluated on ACDC.
Since raindrops and snow are associated with high-frequency components that contain fine details, we plan to further investigate high-frequency components and will improve our method.

{
    \small
    \bibliographystyle{ieeenat_fullname}
    \bibliography{main}
}
\end{document}